\documentclass[10pt,twocolumn,letterpaper]{article}

\usepackage{cvpr}
\usepackage{times}
\usepackage{epsfig}
\usepackage{graphicx}
\usepackage{amsmath}
\usepackage{amssymb}
\usepackage{multirow}

\usepackage[pagebackref=true,breaklinks=true,letterpaper=true,colorlinks,bookmarks=false]{hyperref}

\cvprfinalcopy 

\def\cvprPaperID{****} 

\ifcvprfinal\fi
\begin{document}

\title{Deep Convolutional Generative Adversarial Networks \\Based Flame Detection in Video}

\author{S{\"u}leyman Aslan, U\u{g}ur G{\"u}d{\"u}kbay \\
	Department of Computer Engineering, Bilkent University\\
	Ankara, Turkey\\
	{\tt\small suleyman.aslan@bilkent.edu.tr, gudukbay@cs.bilkent.edu.tr}
	\and
	B. U\u{g}ur T{\"o}reyin 
	\thanks{B.U.~T{\"o}reyin's work is in part funded by T{\"U}B{\.I}TAK 114E426 and {\.I}T{\"U} BAP MGA-2017-40964.}\\
	Informatics Institute, Istanbul Technical University \\
	Istanbul, Turkey\\
	{\tt\small toreyin@itu.edu.tr}
	\and
	A. Enis \c{C}etin 
	\thanks{A.E. \c{C}etin is on leave from Bilkent University and his work is partially funded by NSF with grant number 1739396 and NVIDIA Corporation.}\\
	Dept. of Electrical and Computer Eng., University of Illinois at Chicago \\
	Chicago, IL, USA\\
	{\tt\small aecyy@uic.edu}
}

\maketitle

\begin{abstract}
   Real-time flame detection is crucial in video based surveillance systems. We propose a vision-based method to detect flames using Deep Convolutional Generative Adversarial Neural Networks (DCGANs). Many existing supervised learning approaches using convolutional neural networks do not take temporal information into account and require substantial amount of labeled data. In order to have a robust representation of sequences with and without flame, we propose a two-stage training of a DCGAN exploiting spatio-temporal flame evolution. Our training framework includes the regular training of a DCGAN with real spatio-temporal images, namely, temporal slice images, and noise vectors, and training the discriminator separately using the temporal flame images without the generator. Experimental results show that the proposed method effectively detects flame in video with negligible false positive rates in real-time.
\end{abstract}

\section{Introduction}

\label{sec:intro}
Fires pose great danger in open and large spaces. Flames may spread fast and cause substantial damages to properties and human life. Hence, immediate and accurate flame detection plays instrumental role in fighting fires.

\begin{figure*}[htbp]
	\centering
	\includegraphics[width=0.85\textwidth]{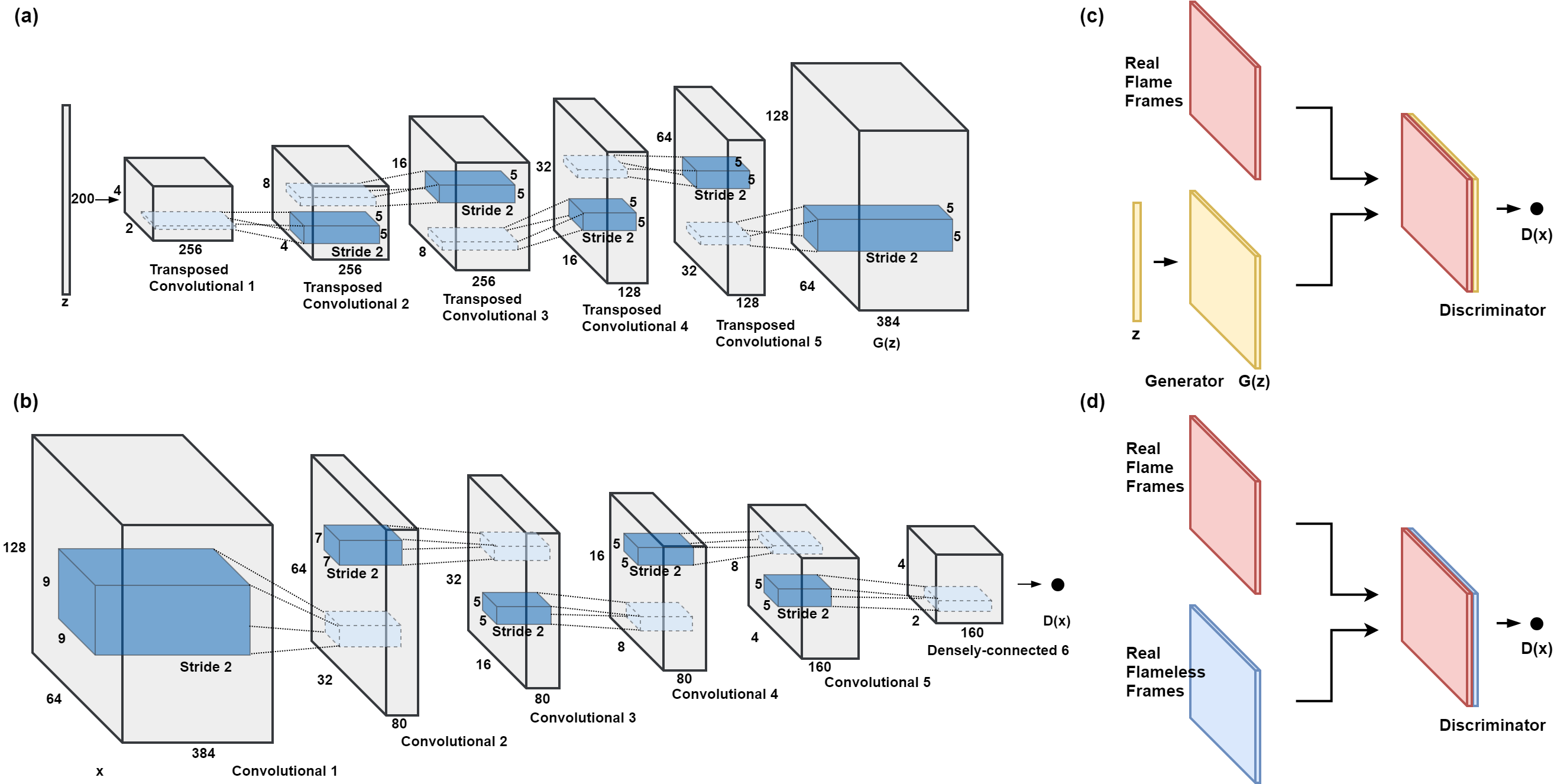}
	\caption{The architecture of DCGAN: (a) generator network,  (b) discriminator network, (c) the first stage of training, and (d) the second stage of training.}	
	\label{fig:flame_full_framework}
\end{figure*}

Among different approaches, the use of visible-range video captured by surveillance cameras are particularly convenient for fire detection, as they can be deployed and operated in a cost-effective manner~\cite{ccetin2013video}. One of the main challenges is to provide a robust vision based detection system with negligible false positive rates, while securing rapid response. If the flames are visible, this may be achieved by analyzing the motion and color clues of a video in wavelet domain~\cite{dedeoglu2005real},~\cite{toreyin2006computer}. Similarly, wavelet based contour analysis~\cite{toreyin2006contour} can be used for detection of possible smoke regions. Modeling various spatio-temporal features such as color and flickering, and dynamic texture analysis~\cite{dimitropoulos2015spatio} have been shown to be able to detect fire, as well. In the literature, there are several computer vision algorithms for smoke and flame detection using wavelets, support vector machines, 
Markov models, region covariance, and co-difference matrices~\cite{ccetin2016methods}. An important number of fire detection algorithms in the literature not only employ spatial information, but also use the temporal information~\cite{ccetin2016methods},~\cite{habibouglu2012covariance},~\cite{toreyin2007online}. 

Deep convolutional neural networks (DCNN) achieve superb recognition results on a wide range of computer vision problems~\cite{goodfellow2014generative},~\cite{lecun2015deep}. Deep neural network based fire detection algorithms using regular cameras have been developed by many researchers in recent years~\cite{gunay2012entropy},~\cite{zhao2018saliency},~\cite{gunay2015real}. As opposed to earlier computer vision based fire detection algorithms, in all of the existing DCNN based methods, temporal nature of flames are not utilized. Instead, flames are recognized from image frames.
In this paper, we utilize the temporal behavior of flames to recognize uncontrolled fires. 
Uncontrolled flames flicker randomly. The bandwidth of spectrum of flame flicker can be as high as 10Hz~\cite{erden2012wavelet}. To detect such behavior, we group the video frames and obtain temporal slice images. We process the temporal slices using deep convolutional networks.

Radford et al.~\cite{radford2015unsupervised} demonstrate that a class of convolutional neural networks, namely, Deep Convolutional Generative Adversarial Networks (DCGANs), can learn general image representations on various image datasets. 

We propose a two-stage training approach for a DCGAN in such a way that the discriminator is utilized to distinguish ordinary image sequences without flame from those with flame. Our contribution is the development of a discriminator network classifying regular images from images with flame. We employ the discriminator network of the DCGAN as a classifier. 

The remainder of the paper is organized as follows. In section~2, the proposed flame detection method is described. Experimental results are presented in Section~3. The paper is concluded in the last section.


\section{Method}
\label{sec:method}

The proposed flame detection method is presented in this section. The method is based on grouping the video frames to obtain temporal slice images and processing the temporal slices using a DCGAN structure accepting input with size 64$\times$128$\times$384~px. We use a densely-connected layer followed by five transposed convolutional layers for the generator, and five convolutional layers with a densely-connected layer for the discriminator. The architecture of DCGAN and the training framework are given in Figure~\ref{fig:flame_full_framework}.

We first train the DCGAN using images that contain flame and noise distribution $z$. The discriminator part of the DCGAN learns a representation for the temporal nature of flames and distinguishes non-flame videos, because those are not in the training set. Then, we refine and retrain the discriminator without generator network, where actual non-flame video images obtained from the cameras constitute the ``generated'' training data and regular flame images correspond to ``real'' data as usual. Compared to a generic CNN structure, training the DCGAN using the flame data, noise vector $z$, and the actual non-flame data makes the recognition system more robust.

In our model, for the training of the networks, we use batch normalization~\cite{ioffe2015batch} and dropout ~\cite{srivastava2014dropout} layers after each layer in the generator network, except the last layer. Similarly, for the discriminator network, apart from the last layer, we add Gaussian noise to the inputs and apply dropout after each layer. Convolution layers in the discriminator are initialized according to the ``MSRA'' initialization~\cite{he2015delving}. Finally, we use the Adam optimizer for stochastic optimization~\cite{kingma2014adam}. The representations of algorithms are supported by TensorFlow system~\cite{abadi2016tensorflow}.

\subsection{Temporal Slice Images}
\label{sec:temporalslice}

\begin{figure*}[htbp]
	\centering
	\includegraphics[width=0.85\textwidth]{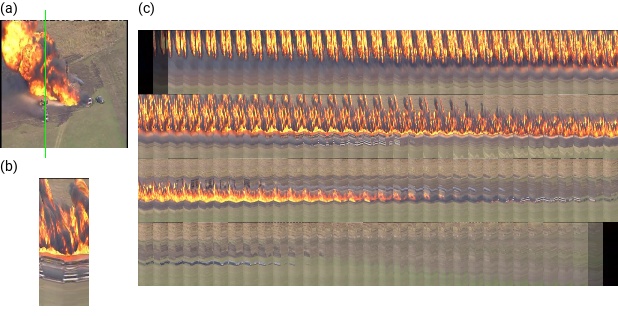}
	\caption{(a) Example frame from the input video. (b) Temporal slice image of column corresponding to the green line in (a), where the leftmost column contains pixels from the initial frame, namely, the frame at time index $t=1$, and the rightmost column contains pixels from the final frame, namely, the frame at time index $t=64$ of the block. (c) Visualization of all 128 slice images.}	
	\label{fig:temporalsliceexamples}	
\end{figure*}

Exploiting the evolution of flames in time, we obtain slice images from video frames. We first split the videos into blocks containing 64 consecutive frames with size 128$\times$128~px. Then, for each column, we extract the pixels along the time dimension, resulting in 128 different 128$\times$64~px images (see Figure~\ref{fig:temporalsliceexamples}).

\begin{figure}[htbp]
	\centering
	\includegraphics[width=8.5cm]{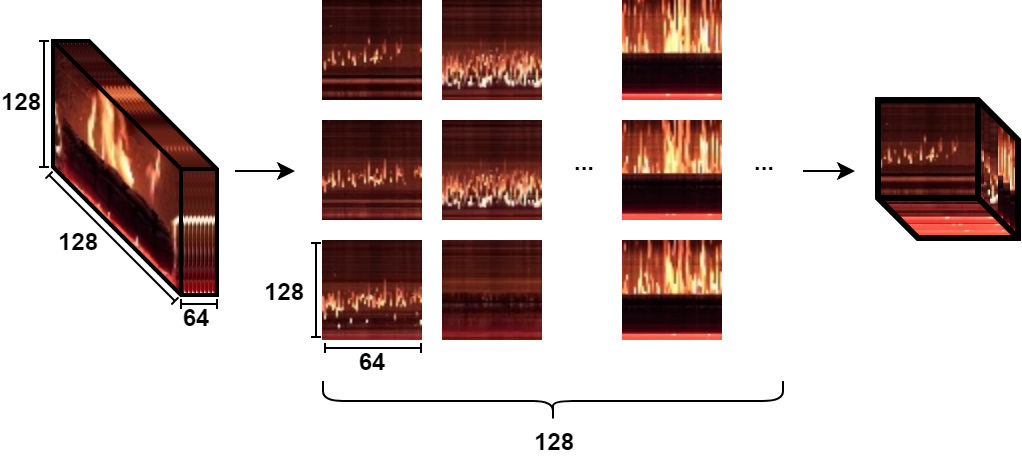}
	\caption{Example of an image cube obtained from the input video.}	
	\label{fig:temporaldata}	
\end{figure}

In order to feed the slice image data to the DCGAN model, we stack all 128 slices on top of each other. Thus, we obtain an RGB image cube of shape 64$\times$128$\times$384, because slice images have 3 channels each. Figure~\ref{fig:temporaldata} shows an example of an image cube. 

\subsection{Proposed GAN-type Discriminator Network}
\label{sec:format}

Flames, by their nature, has no particular shape or specific feature as human faces, cars, and so on. Therefore, it is more suitable to focus on the temporal behavior of flame instead of the spatial information. 

The DCGAN structure is utilized to distinguish regular camera views from flame videos. The discriminator part of the GAN produces probability values above 0.5 for real temporal flame slices and below 0.5 for slices that do not contain flame, because non-flame slices are not in the initial training set. In the second stage of training, we refine and retrain the GAN using the gradient given in \eqref{refine}.

In standard GAN training, the discriminator $D$ which outputs a probability value is updated using the stochastic gradient

\begin{equation}
SG_1= \nabla_{\theta_d}  \frac{1}{M} \sum_{i=1}^M (\log D (x_i) + \log (1-D(G(z_i)))),
\end{equation}

\noindent where $x_i$ and $z_i$ are the $i$-th temporal slice and noise vector, respectively, and $G$ represents the generator which generates a "fake slice" according to the input noise vector $z_i$; the vector $\theta_d$ contains the parameters of the discriminator. After this stage, the generator network $G$ is ``adversarially'' trained, as in~\cite{goodfellow2014generative}. During the first round of training we do not include any flame-less video. This GAN is able to distinguish flame, because regular camera views are not in the training set. To increase the recognition accuracy, we perform a second round of training by fine-tuning the discriminator using the stochastic gradient 

\begin{equation}
SG_2 = \nabla_{\theta_d} \frac{1}{L} \sum_{i=1}^L (\log D (x_i) + \log (1-D(y_i)),
\label{refine}
\end{equation}

\noindent where $y_i$ represents the $i$-th image containing regular camera views. The number of non-flame slice samples, $L$, is smaller than the size of the initial training set, $M$, containing flame videos. In the refinement stage characterized by \eqref{refine},
we do not update the parameters of the generator network of GAN, because we do not need to generate any artificial images at this stage of training.

\section{Experimental Results}
\label{sec:experimentalresults}

In our experiments, we use 112 video clips containing flame frames, and 72 video clips without any flame frames. Flame videos contain various events, such as burning buildings, fire explosions, fireplaces, campfires, forest fires, and burning vehicles.

Throughout the experiments, we first obtain the temporal slice images for both flame and non-flame videos. For that purpose, at every second, we sample 10 previous frames at equal intervals, to be included in a block. Since blocks contain 64 frames, they capture the motion for almost six and a half seconds. Video clips are partitioned into non-overlapping temporal slices. Each video clip has a duration of one minute. Consequently, the dataset is composed of over 210 thousand slices from over 1600 blocks in total. 

After this procedure, we split the data into training, validation, and test sets with a ratio of 3:1:1. We pick the parameters and stop training the network based on its performance on the validation set, then report the final results obtained on the test set.

We evaluate the proposed method, namely, DCGAN with Temporal
Slices, in terms of frame-based results. Since all the other deep learning methods are essentially based on CNNs, we compare the CNN with Temporal Slices, DCGAN with Video Frames (no temporal information) and DCGAN without refinement stage based approaches to our CNN implementation. It should be also noted that, researchers use different fire datasets, therefore the recognition results are not comparable.

Our approach targets at reducing the false positive rate, while keeping the hit-rate, as high as possible. Results indicate that, our methode achieves the best results on the test set (cf. Table~\ref{table:framebasedcomparison}), where a false-positive rate of 3.91\% is obtained corresponding to a hit-rate of 92.19\%. We show that the adversarial training in DCGAN structure yields more robust results when compared to a CNN (same architecture as the discriminator). As for the utilization of temporal slices to exploit flame evolution, it can be seen that, utilizing the temporal information of the flames results in much lower false positive rates. 

\begin{figure*}[htb]
\centering
\vspace{0.8cm}
\includegraphics[width=0.85\textwidth]{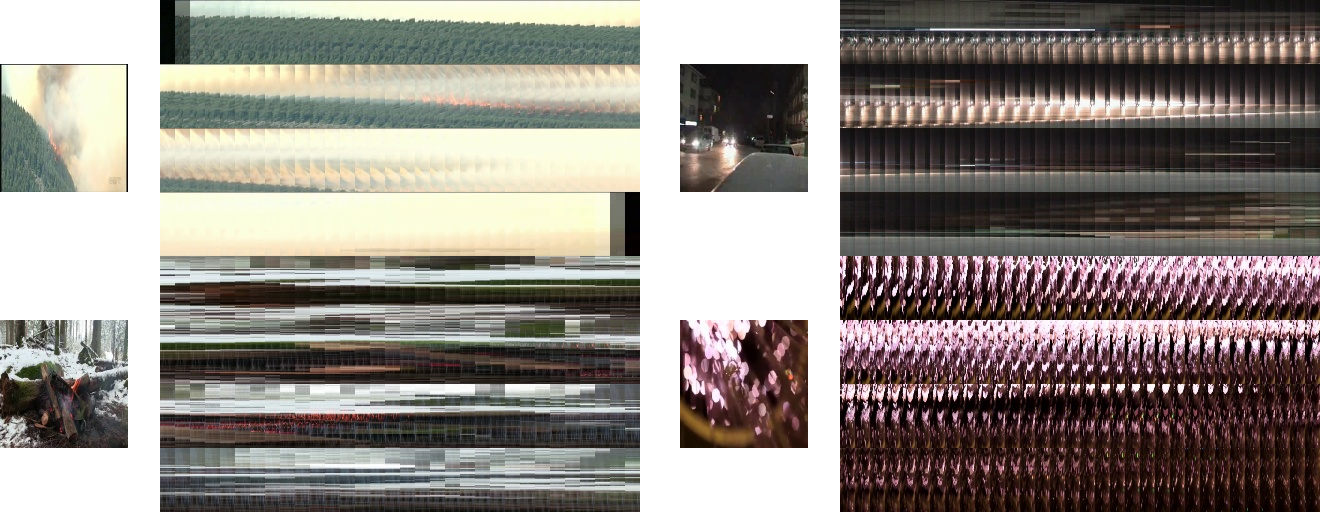}
\caption{Examples of false negative temporal slices on the left and false positive temporal slices on the right.}
\label{fig:falseclassify}
\end{figure*}

Some examples for false negative and false positive temporal slices are presented in Figure~\ref{fig:falseclassify}.

\begin{table}[htb]
	\centering
	\caption{Obtained true negative rate (TNR) and true positive rate (TPR) values on test set for frame-based evaluation.}
	\label{table:framebasedcomparison}
	\begin{tabular}{|p{4.2cm}|c|c|}
		\hline
		\textbf{Method}&\textbf{TNR}&\textbf{TPR}\\ 
		\textbf{}&\textbf{(\%)}&\textbf{(\%)}\\ \hline
		DCGAN with Temporal Slices (Our method)&96.09&92.19\\ \hline
		CNN with Temporal Slices&87.39&93.23\\ \hline
		DCGAN with Video Frames (no temporal information)&92.55&92.39\\ \hline
		DCGAN without refinement stage&86.61&90.10\\ \hline
	\end{tabular}
\end{table}

\section{Conclusion}
\label{sec:conclusion}
We propose a fire detection method using DCGANs exploiting spatio-temporal evolution of flames. We develop a two-stage DCGAN training approach in order to classify flame and non-flame image sequences. Spatio-temporal dynamics of flames are acquired using temporal slice images obtained from consecutive frames. 

Results suggest that the proposed method achieves low false alarm rates while keeping the detection rate high, as opposed to the other deep learning approaches.

{\small
\bibliographystyle{ieee}
\bibliography{refs}
}

\end{document}